\documentclass[10pt,twocolumn,letterpaper]{article}

\usepackage{cvpr}              

\usepackage{graphicx}
\usepackage{amsmath}
\usepackage{amssymb}
\usepackage{booktabs}

\usepackage{multirow} 
\usepackage{array}    

\usepackage[pagebackref,breaklinks,colorlinks]{hyperref}

\usepackage[capitalize]{cleveref}
\crefname{section}{Sec.}{Secs.}
\Crefname{section}{Section}{Sections}
\Crefname{table}{Table}{Tables}
\crefname{table}{Tab.}{Tabs.}

\def\cvprPaperID{2238} 
\def\confName{CVPR}
\def\confYear{2023}

\begin{document}

\title{
A Unified HDR Imaging Method with Pixel and Patch Level
}

\author{Qingsen Yan$^\dag$$^1$, Weiye Chen$^\dag$$^2$, Song Zhang$^\dag$$^2$, Yu Zhu$^1$, Jinqiu Sun$^1$, Yanning Zhang$^1$
\thanks{$\dag$~The first three authors contributed equally to this work.
This work was partially supported by NSFC (U19B2037, 61901384), Natural Science Basic Research Program of Shaanxi (2021JCW-03, 2023-JC-QN-0685), the Fundamental Research Funds for the Central Universities (D5000220444), and National Engineering Laboratory for Integrated Aero-Space-Ground-Ocean Big Data Application Technology. Corresponding author: Jinqiu Sun.}\\
$^1$Northwestern Polytechnical University,
$^2$Xidian University
}

\maketitle

\begin{abstract}

Mapping Low Dynamic Range (LDR) images with different exposures to High Dynamic Range (HDR) remains nontrivial and challenging on dynamic scenes due to ghosting caused by object motion or camera jitting. With the success of Deep Neural Networks (DNNs), several DNNs-based methods have been proposed to alleviate ghosting, they cannot generate approving results when motion and saturation occur. To generate visually pleasing HDR images in various cases, we propose a hybrid HDR deghosting network, called HyHDRNet, to learn the complicated relationship between reference and non-reference images. The proposed HyHDRNet consists of a content alignment subnetwork and a Transformer-based fusion subnetwork. Specifically, to effectively avoid ghosting from the source, the content alignment subnetwork uses patch aggregation and ghost attention to integrate similar content from other non-reference images with patch level and suppress undesired components with pixel level. To achieve mutual guidance between patch-level and pixel-level, we leverage a gating module to sufficiently swap useful information both in ghosted and saturated regions. Furthermore, to obtain a high-quality HDR image, the Transformer-based fusion subnetwork uses a Residual Deformable Transformer Block (RDTB) to adaptively merge information for different exposed regions. We examined the proposed method on four widely used public HDR image deghosting datasets. Experiments demonstrate that HyHDRNet outperforms state-of-the-art methods both quantitatively and qualitatively, achieving appealing HDR visualization with unified textures and colors.
\end{abstract}

\section{Introduction}
Natural scenes cover a very broad range of illumination, but standard digital camera sensors can only measure a limited dynamic range. Images captured by cameras often have saturated or under-exposed regions, which lead to terrible visual effects due to severely missing details. High Dynamic Range (HDR) imaging has been developed to address these limitations, and it can display richer details. A common way of HDR imaging is to fuse a series of differently exposed Low Dynamic Range (LDR) images. It can recover a high-quality HDR image when both the scene and the camera are static, however, it suffers from ghosting artifacts on dynamic objects or hand-held camera scenarios.\par

\begin{figure}[t!]
    \centering
    \includegraphics[width=0.45\textwidth]{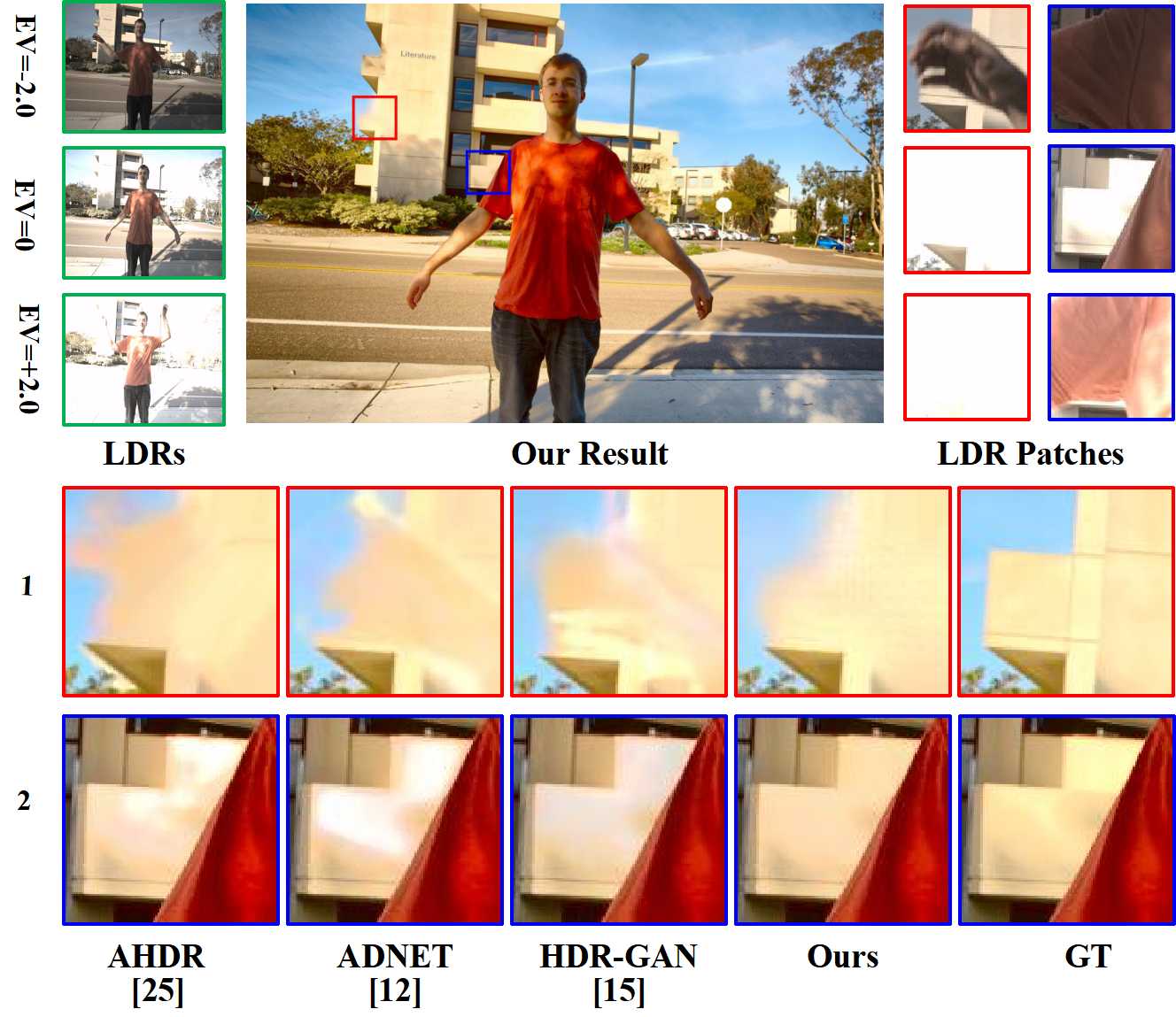}
    \caption{Our approach produces high-quality HDR images, leveraging both patch-wise aggregation and pixel-wise ghost attention. The two modules provide complementary visual information: patch aggregation recovers patch-level content of the complex distorted regions and ghost attention provides pixel-level alignment.}
    \label{fig:intro}
    \vspace{-0.5cm}
\end{figure}

Several methods have been proposed to alleviate these problems, including alignment-based methods \cite{Bogoni2000Extending, KangSig2003High, Tomaszewska2007Image}, rejection-based methods \cite{Grosch2006Fast, Pece2010Bitmap, Zhang2011Gradient, Lee2014Ghost, Oh2015Robust} and patch-based methods \cite{Sen2012Robust, Hu2013HDR, Ma2017Robust}. The alignment-based methods employ global (\eg, homographies) or non-rigid alignment (\eg, optical flow) to rectify the content of motion regions, but they are error-prone and cause ghosts due to saturation and occlusion. The rejection-based methods attempt to remove the motion components from the exposed images and replace them with the content of the reference image. Although these methods achieve good quality for static scenes, they discard the misalignment regions, which causes insufficient content in moving regions. The patch-based methods utilize patch-level alignment to transfer and fuse similar content. While these patch-based methods achieve better performance, they suffer from high computational costs.


With the rise of Deep Neural Networks (DNNs), many works directly learn the complicated mapping between LDR and HDR using a CNN. In general, these models follow the paradigm of alignment before fusion. The non-end-to-end DNN-based methods \cite{Kalantari2017Deep,Wu2018Deep} first align LDR images with optical flow or homographies, and then fuse aligned images to generate HDR images. The alignment approaches are error-prone and inevitably cause ghosting artifacts when complex foreground motions occur.
Based on the attention-based end-to-end method AHDRNet \cite{Yan2019Attention,yan2022dual} which performs spatial attention to suppress motion and saturation, several methods \cite{Liu2021ADNet, yan2022high, Chen2022Attention, yan2020ghost,Yan2022Alightweight} have been proposed to remove ghosting artifacts. The spatial attention module produces attention maps and element-wise multiplies with non-reference features, thus the model removes motion or saturated regions and highlights more informative regions.

However, the success of these methods relies on noticeable variations between reference and non-reference frames. These methods perform well in the marginal areas, even if there is a misalignment in the input images. Unluckily, spatial attention produces unsatisfactory results when motion and saturation are present simultaneously (see Figure \ref{fig:intro}). The reason can be attributed to that spatial attention uses element-wise multiplication, which only considers the features in the same positions. For example, in the reference frame of Figure \ref{fig:intro} (\ie, LDR with EV=0), the information in the over-exposed regions is unavailable, spatial attention can only rely on the non-saturated information of the same position (\ie, moving regions) in the non-reference frame due to element-wise multiplication. Therefore, recovering the content of the moving and saturated regions is challenging. Finally, this limitation of spatial attention causes obvious ghosting artifacts in these complex cases. 

To generate high-quality HDR images in various cases, we propose a Hybrid HDR deghosting Network, named HyHDRNet, to establish the complicated alignment and fusion relationship between reference and non-reference images. The proposed HyHDRNet comprises a content alignment subnetwork and a Transformer-based fusion subnetwork. For the content alignment subnetwork, inspired by patch-based HDR imaging methods \cite{Sen2012Robust, Hu2013HDR}, we propose a novel Patch Aggregation (PA) module, which calculates the similarity map between different patches and selectively aggregates useful information from non-reference LDR images, to remove ghosts and generate content of saturation and misalignment. 
While the traditional patch-based HDR imaging methods have excellent performance but have the following drawbacks:
1) low patch utilization ratio caused by reusing the same patches, which leads to insufficient content during fusion,
2) structural destruction of images when transfering patches,
3) high computational complexity in full resolution.
To this end, our Patch Aggregation mechanism
1) aggregates multiple patches which improves the patch utilization ratio 
2) aggregates patches instead of exchanging them to maintain structural information,
3) calculates a similarity map within a window to reduce computational complexity.
These advantages promote the network to remedy the content of saturated and motion regions(See Figure \ref{fig:supp_ablation2}), other patch-based HDR imaging methods cannot achieve this goal.
In a word, our PA module (patch level) discovers and aggregates similar patches within a large receptive field according to the similarity map, thus it can recover the content inside the distorted regions.
To further avoid ghosting, we also employ a ghost attention module (pixel level) as a complementary branch for the PA module, and propose a gating module to achieve mutual guidance of these two modules in the content alignment subnetwork. In addition, unlike previous methods using DNN structure in the feature fusion stage which has static weights and only merges the local information, we propose a Transformer-based fusion subnetwork that uses Residual Deformable Transformer Block (RDTB) to model long-range dependencies of different regions. The RDTB can dynamically adjust weights and adaptively merge information in different exposure regions. The experiments demonstrate that our proposed method achieves state-of-the-art performance on public datasets. The main contributions of our work can be summarized as follows:\par

\begin{itemize}
    \vspace{-0.3cm}
    \item We propose a hybrid HDR deghosting network to effectively integrate the advantage of patch aggregation and ghost attention using a gating strategy. 
    \vspace{-0.2cm}
    \item We first introduce the patch aggregation module which selectively aggregates useful information from non-reference LDR images to remove ghosts and generate content for saturation.
    \vspace{-0.2cm}
    \item A novel residual deformable Transformer block is proposed, which can adaptively fuse a large range of information to generate high-quality HDR images.
    \vspace{-0.2cm}
    \item We carry out both qualitative and quantitative experiments, which show that our method achieves state-of-the-art results over four public benchmarks.
\end{itemize}

\begin{figure*}[t]
    \centering
    \includegraphics[width=1\textwidth]{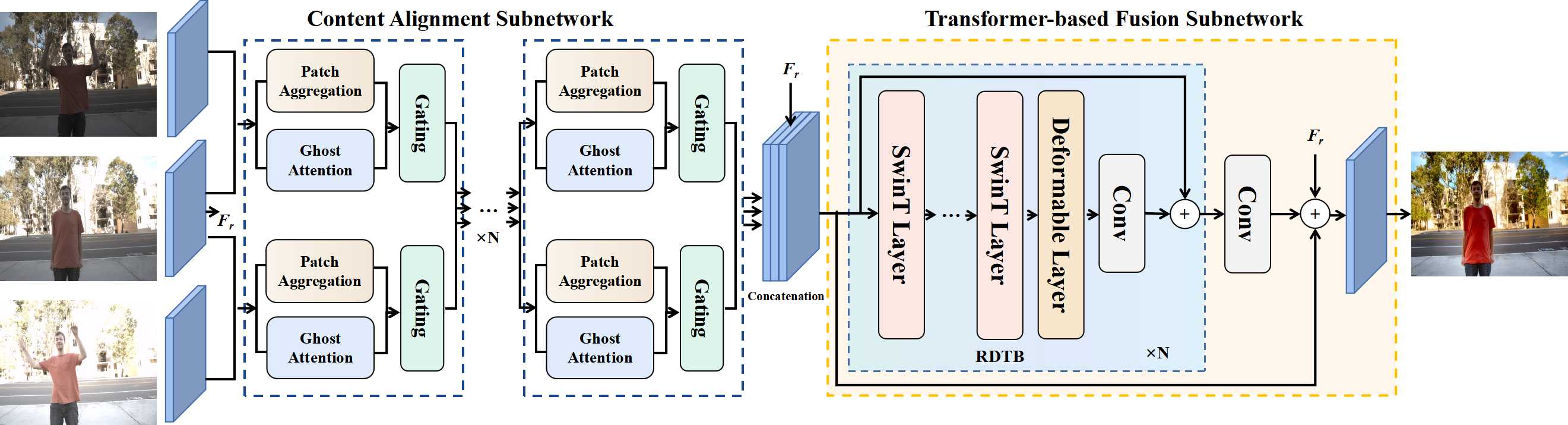}
    \caption{The illustration of HyHDRNet which consists of a content alignment subnetwork and a Transformer-based fusion subnetwork. The content alignment subnetwork uses patch aggregation and ghost attention to integrate similar content from other LDR images with patch level and suppress undesired components with pixel level. A gating module is leveraged to achieve mutual guidance between patch-level and pixel-level. Then the Transformer-based fusion subnetwork uses Residual Deformable Transformer Block (RDTB) to adaptively merge information for different exposed regions to obtain a high-quality HDR image.}
    \label{fig:overview}
\end{figure*}

\vspace{-0.4cm}
\section{Related work}
The related work includes four categories: alignment-based method, rejection-based method, patch-based method and CNN-based method.

\textbf{Alignment-based method.}
Alignment methods utilize rigid or non-rigid registration to match reference images densely. Bogoni \etal \cite{Bogoni2000Extending} computed flow vectors for alignment and used pattern selective fusion to obtain HDR images. Kang \etal \cite{KangSig2003High} aligned images of video using optical flow in the luminance domain and fused the aligned images to remove ghosting artifacts. Tomaszewska \etal \cite{Tomaszewska2007Image} employed SIFT feature to eliminate global misalignments, which can reduce blurring and artifacts. Although alignment methods can find dense correspondence, they are easy to fail in large motion and occlusion regions.

\textbf{Rejection-based method.}
Rejection methods identify and remove motion regions of inputs after global registration operation, then fuse static regions to reconstruct HDR images. Grosch \etal \cite{Grosch2006Fast} marked motion regions and used the predicted color's error map to obtain ghost-free HDR. Pece \etal \cite{Pece2010Bitmap} employed the median threshold bitmap of the input image to reject motion regions. Zhang \etal \cite{Zhang2011Gradient} used quality measures based on image gradients to generate a motion weighting map. Several methods \cite{Lee2014Ghost,Oh2015Robust,yan2017high} utilized rank minimization to detect the moving region and reconstruct ghost-free HDR images. However, pixel rejection abandons the misaligned regions, which causes insufficient content in moving regions.

\textbf{Patch-based method.}
The patch-based methods use patch-wise alignment between the exposure images for deghosting. Sen \etal \cite{Sen2012Robust} proposed a patch-based energy minimization method, and it optimizes alignment and reconstruction jointly. Hu \etal \cite{Hu2013HDR} propagated intensity and gradient information iteratively using a coarse-to-fine schedule. Ma \etal \cite{Ma2017Robust} proposed a structural patch decomposition approach that decomposes an image patch into signal strength, signal structure and mean intensity components to reconstruct ghost-free images. But these methods do not compensate for saturation and suffer from high computational costs.

\textbf{CNN-based method.}
Kalantari \etal \cite{Kalantari2017Deep} used a convolutional neural network to fuse LDR images after optical flow alignment. Wu \etal \cite{Wu2018Deep} defined HDR imaging as an image translation problem, and used an image homography transformer to align the camera motion.
Yan \etal \cite{Yan2019Attention} leveraged an attention mechanism to suppress undesired information to merge ghost-free HDR. Yan \etal \cite{Yan2020Deep} designed a nonlocal block to learn the constraint of locality receptive field for global merging HDR. Niu \etal \cite{Niu2021Hdr} proposed HDR-GAN to synthesize missing content using Generative Adversarial Networks. 
Xiong \etal \cite{Xiong2021Hierarchical} formulated HDR imaging as two problems, ghost-free fusion and ghost-based restoration. Ye \etal \cite{Ye2021Progressive} proposed multi-step feature fusion and comparing/selecting operations to generate ghost-free images. Liu \etal \cite{Liu2021ADNet} utilized the PCD pyramid alignment subnetwork to register in multiple layers.
These methods still use pixel-level information to complete HDR imaging, which neglects the content of the patch level.

\section{Proposed method}
\subsection{Problem Definition}
Given a sequence of LDR images $\{L_{1},L_{2},......,L_{N}\}$ with different exposures in dynamic scenes, our goal is to merge them into an HDR image without ghosting artifacts. The estimated HDR image will be structurally similar to a reference image $L_{r}$. Following previous works \cite{Kalantari2017Deep,Yan2019Attention}, we utilize three LDR images $(L_{1},L_{2},L_{3})$ as input, and set $L_{2}$ as reference image.\par

Following \cite{Kalantari2017Deep, Yan2019Attention}, we first map the LDR images $L_{i}$ to the HDR domain using gamma correction:
\begin{equation}
    H_i = L_i^{\gamma} / t_i,
\end{equation}
where $t_i$ denotes the exposure time of LDR image $L_i$, $\gamma$ represents the gamma correction parameter, we set $\gamma$ to 2.2. Then we concatenate $L_i$ and $H_i$ along the channel dimension to get a 6-channel input $X_i=[L_i, H_i]$. Given inputs $X_1$, $X_2$, $X_3$, our model produces an HDR image $\hat H$ by:
\begin{equation}
    \hat{H} = f(X_1,X_2,X_3;\theta),
\end{equation}
where $f(\cdot)$ represents the HDR imaging network, $\theta$ is the parameters of the network.

\subsection{Overview}
As shown in Figure \ref{fig:overview}, the proposed HyHDRNet consists of two subnetworks, a content alignment subnetwork and a Transformer-based fusion subnetwork. 
The content alignment subnetwork first extracts shallow features from LDR sequence using three individual convolutional layers. Then the features utilize Patch Aggregation (PA) and Ghost Attention (GA) modules to separately aggregate useful patches from non-reference LDR images for complex cases and identify the misaligned components for easy cases. To take full advantage of these two modules, we use the gating module to recover both contents of saturated regions and sharp edges in motion and saturation. Considering that the missed content can be filled up using the information of neighborhood regions from the non-reference image, the PA module can search and assemble similar patches to complete misaligned or saturated regions, and the GA module is a complementary branch that removes ghosting with pixel level.

The Transformer-based fusion subnetwork aims to generate high-quality ghost-free HDR images from the extracted features of LDR inputs. We first utilize the features extracted from the content alignment subnetwork as input, then use several Swin Transformer layers and deformable layers to model long-range dependencies and dynamically merge useful information with a larger receptive field. Finally, we adopt convolutional layers to obtain a 3-channel HDR image.



\subsection{Content Alignment Subnetwork}

\textbf{Shallow Feature Extraction Module.}
Given three 6-channel LDR images $X_{i} \in R^{H \times W \times 6}$, $i=$1, 2, 3, we adopt three convolutional encoders to extract the shallow feature $F_{i} \in R^{H \times W \times C}$:
\begin{equation}
    F_{i}=e(X_{i}), i=1,2,3
\end{equation}

\begin{figure}
    \centering
    \includegraphics[width=0.45\textwidth, height=0.12\textwidth]{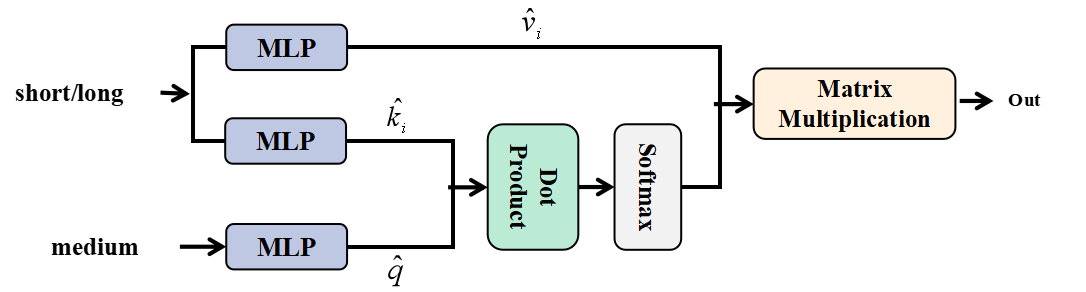}
    \caption{The structure of patch aggregation module. The PA module calculates the similarity map between different patches and selectively aggregates useful information from non-reference LDR images to remove ghosts and generate content of saturation and misalignment.}
    \label{fig:pa}
    \vspace{-0.3cm}
\end{figure}

\begin{figure}
    \centering
    \includegraphics[width=0.45\textwidth, height=0.12\textwidth]{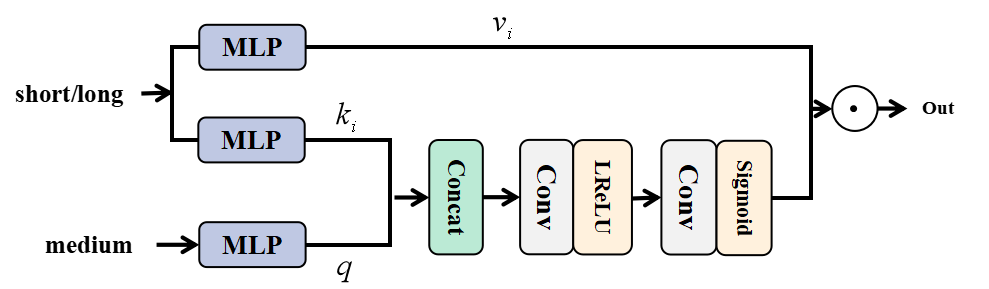}
    \caption{The structure of ghost attention module. The GA module utilizes spatial attention to identify the misaligned components.}
    \label{fig:sa}
\end{figure}

\textbf{Patch Aggregation Module.}
Traditional patch-based HDR imaging methods have excellent performance, but are limited by a low patch utilization ratio which causes insufficient content during fusion and has high computational complexity in full resolution.


In order to overcome the shortcoming of element-wise multiplication in spatial attention, the PA module selectively aggregates information by calculating the similarity map between patches. As shown in Figure \ref{fig:pa}, we first divide the shallow feature $F_{i} \in R^{R \times W \times C}$ into ${HW}/{M^{2}}$ non-overlapping $M \times M$ patches $\hat F_{i}$. We map each patch $\hat F_{i}$ to query $\hat q$, keys $\hat k_{i}$ and values $\hat v_{i}$:
\begin{equation}
    \hat{q}=\hat{F_{r}}W_{q},\hat{k_{i}}=\hat{F_{i}}W_{k},\hat{v_{i}}=\hat{F_{i}}W_{v},i=1,3
\end{equation}
where $\hat{F_{r}}$ is the features of the reference image.

Then we calculate the similarity map between non-reference feature $\hat k_{i}$ and reference feature $\hat q$, and apply the softmax function to the similarity map. Finally, we aggregate patches from non-reference feature $\hat v_{i}$ according to the similarity map, and obtain the output feature $F_{pa}^i$.
\begin{equation}
    F_{pa}^{i}=Softmax(\hat{q} \hat{k_{i}^{T}} / \sqrt{d} + B)\hat{v_{i}},
\end{equation}
where $B$ is a learnable position encoding, $d$ is the dimension of $\hat q$. 
Note that, we also leverage the shifted windows \cite{liu2021swin} to achieve interaction of each patch.

\textbf{Ghost Attention Module.}
Spatial Attention \cite{Yan2019Attention} can suppress motion and saturation well at the pixel level. As shown in Figure \ref{fig:sa}, we design a ghost attention which first maps the shallow feature $F_i$ to query, keys and values. $F_r$ is mapped from $X_2$, query $q$ is mapped from $F_r$, keys $k_{i}$ and values $v_{i}$ are mapped from non-reference frame $F_i$, $i=$1,3. Note that $W_{q}$, $W_{k}$, $W_{v}$ are shared in PA and GA.
\begin{equation}
    q=F_{r}W_{q},k_{i}=F_{i}W_{k},v_{i}=F_{i}W_{v},i=1,3
\end{equation}
Then, spatial attention $a(\cdot)$ calculate the attention between the non-reference feature $k_{i}$ and reference feature $q$.
\begin{equation}
    A_{i}=a(q, k_{i}), i=1,3
\end{equation}
Finally, we obtain the output feature $F_{ga}$ by element-wise multiplication of non-reference feature $v_{i}$ and attention map $A_i$, $\odot$ represents element-wise multiplication.
\begin{equation}
    F_{ga}^{i} = v_i \odot A_i, i=1,3
\end{equation}


\begin{figure}[t]
    \centering
    \includegraphics[width=0.4\textwidth]{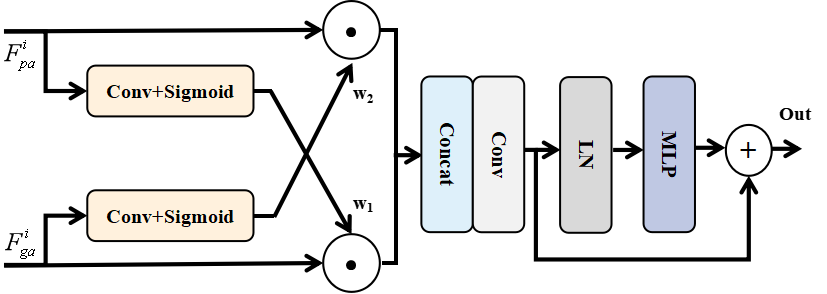}
    \caption{The structure of gating module. The gating module achieves mutual guidance of PA and GA modules in the content alignment subnetwork.}
    \label{fig:gating}
\end{figure}

\textbf{Gating Module.}
In order to integrate the advantage of PA and GA modules, we propose a gating module to achieve mutual guidance between these two modules. As shown in Figure \ref{fig:gating}, given the output features $F_{pa}$ and $F_{ga}$ from the PA module and GA module, we use $\varphi$ function to predict the weight $w_1$ and $w_2$ for $F_{ga}$ and $F_{pa}$, respectively. Then the element-wise multiplicated features are concatenated as inputs of a convolutional layer. Then we adopt residual connection, LN layer and MLP layer to obtain the output $F_{out}$, denoted by:
\begin{equation}
    F_{out} = F_{gating} + LN(MLP(F_{gating})),
\end{equation}
\begin{equation}
    F_{gating} = Conv(Concat(F_{ga} \odot \varphi(F_{pa}), F_{pa} \odot \varphi(F_{ga})))),
\end{equation}
where $LN(\cdot)$ is a layernorm layer, $\varphi$ consists of a convolution and a sigmoid function. 



\subsection{Transformer-based Fusion Subnetwork}
Inspired by SWIN-IR \cite{Liang2021Swinir} and Deformable Attention Transformer \cite{Xia2022Vision}, we use several Residual Deformable Transformer Blocks (RDTB) to dynamically merge features for generating high-quality HDR image. Compared with \cite{Xia2022Vision}, we design a Window-based Deformable Transformer Layer (WDTL) to maintain the performance and decrease the computational cost, simultaneously. In addition, we also utilize channel attention in FFN to make the WDTL block converge more quickly and smooth the loss landscape.

\begin{figure}
    \centering
    \includegraphics[width=0.5\textwidth]{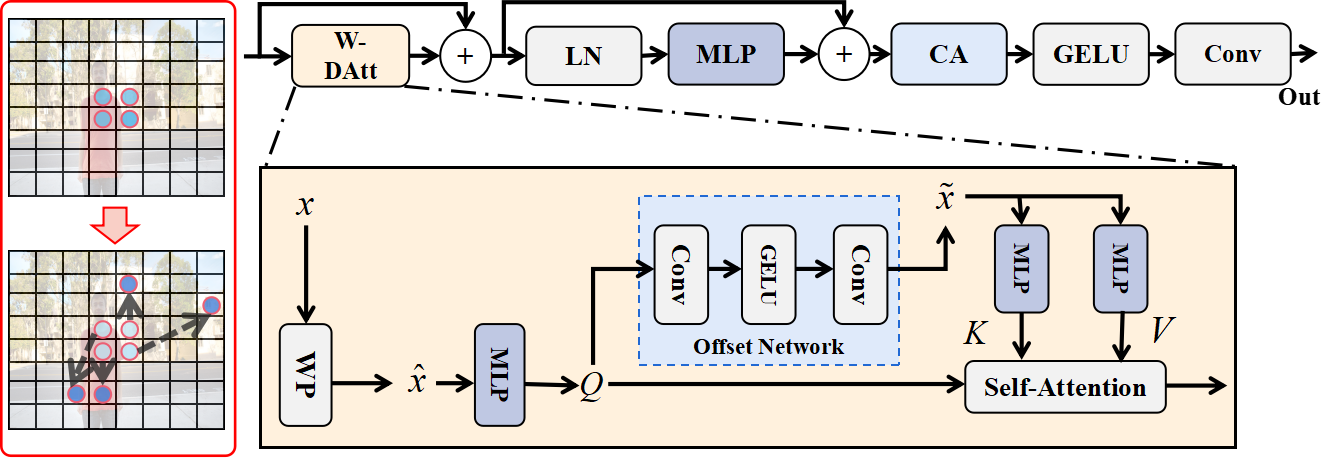}
    \caption{The window-based deformable Transformer layer. We uniformly place a group of reference points on the feature map, and the offsets of these reference points are learned from the query by the offset network. Then the deformed keys and values are projected from the sampled features with the deformed points, the transformed features are obtained by the self-attention block.
}
    \label{fig:dfl}
    \vspace{-0.3cm}
\end{figure}

\textbf{Residual Deformable Transformer Block.}
As shown in Figure \ref{fig:overview}, our RDTB consists of Swin Transformer Layers (STL), Window-based Deformable Transformer Layers (WDTL), convolutional layers and residual connection. Given the input feature $F_{i}^{0}$ of $i$-th RDTB, the output of RDTB can be formulated as:
\begin{equation}
    R_{out}=Conv(WDTL(STL_{i}^{N}))+F_{i}^{0},
\end{equation}
where $WDTL(\cdot)$ denotes Deformable Layer, $STL_{i}^{N}(\cdot)$ denotes the output of $N$-th Swin Transformer layer in $i$-th RDTB.
STL includes multi-head intra-/inter-window self-attention and Feed-Forward Network (FFN), more details can be found in \cite{Liang2021Swinir}. We will describe the details of WDTL in the following.

\textit{Window-based Deformable Transformer Layer.}
As shown in Figure \ref{fig:dfl}, given input feature $x$, we first put $x$ into Window-based Deformable self-Attention (WDAtt). We use Window Partition (WP) to divide the input into local windows $\hat x$, then we map them to query embedding $Q$ with $W'_{q}$. Afterwards, we use a lightweight network $O_{offset}(\cdot)$ which consists of two convolutional layers and an activation function to predict the offsets $\Delta o$ of each window, as shown in Figure \ref{fig:dfl}.
We then get the deformable windows $\tilde{x}$ and map them to embedding $K$ and $V$. 
\begin{equation}
    Q=\hat{x}W'_{q}, \Delta o = O_{offset}(Q), \tilde{x} = \Phi(\hat{x};p+\Delta o),
\end{equation}
\begin{equation}
   K=\tilde{x}W'_{k}, V=\tilde{x}W'_{v}.
\end{equation}
Finally, we obtain transformed features $X_{tr}$ using $Q$, $K$, $V$.
\begin{equation}
    X_{tr}=Softmax(QK^{T}/\sqrt{D})V,
\end{equation}
where $\Phi$ denotes the bilinear interpolation function, $D$ represents the dimension of $Q$.

In order to make the WDTL block converge more easily and make the loss landscape more smoothing, inspired by \cite{Namuk2022How}, we utilize channel attention in the FFN layer to get $X_{DT}$, denoted by:
\begin{equation}
    X_{tr}^{'}=MLP(LN(X_{tr}))+X_{tr},
\end{equation}
\begin{equation}
    X_{DT}=Conv(GELU(CA(X_{tr}^{'}))),
\end{equation}
where $CA(\cdot)$ is the channel attention block, $GELU(\cdot)$ denotes the GELU non-linearity function.

\subsection{Training Loss}
Since the HDR images are displayed in the tonemapped domain, we use $\mu$-law \cite{Kalantari2017Deep} to map the image from the linear domain to the tonemapped domain:
\begin{equation}
    T(x)=\frac{log(1+\mu x)}{log(1+\mu)},
\end{equation}
where $T(x)$ is the tonemapped function, $\mu=5000$. 


Given the estimated result $\hat{H}$ of HyHDRNet and the ground truth $H$, we calculate the tonemapped per-pixel loss (first term) and perceptual loss (second term) as follows:
\begin{equation}
    L_{total} = ||T(H)-T(\hat{H}))||_1 + \lambda ||\phi_{i,j}(T(H))-\phi_{i,j}(T(\hat{H}))||_1
\end{equation}
where $\phi_{i,j}$ represents the $j$-th convolutional feature extracted from VGG19 after the $i$-th max-pooling operation, $\lambda=1e^{-2}$ is a weighting hyperparameter.

\section{Experiments}

\begin{figure*}[t!]
  \centering
    \includegraphics[width=0.98\textwidth]{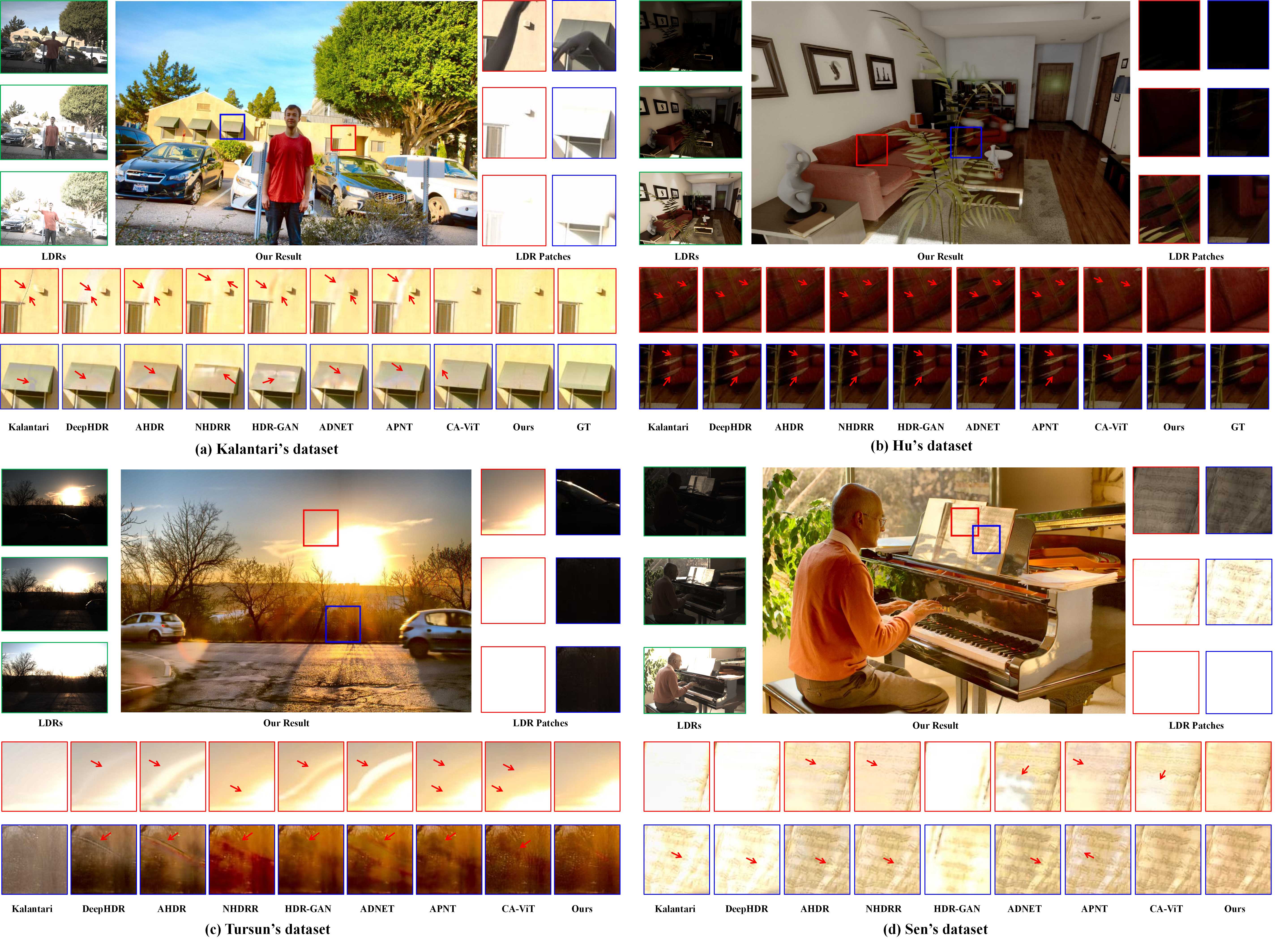}
  \caption{Examples of Kalantari’s dataset \cite{Kalantari2017Deep} and Hu’s dataset \cite{Hu2020Sensor} (top row) and Tursun’s dataset \cite{Tursun2016An} and Sen’s dataset \cite{Sen2012Robust} (bottom row) datasets.}
  \label{fig:KalanHuTursunSen}
\end{figure*}

\subsection{Experimental Settings}
\textbf{Datasets.}
All methods are trained on two public datasets, Kalantari’s dataset \cite{Kalantari2017Deep} and Hu’s dataset \cite{Hu2020Sensor}. Kalantari’s dataset is captured in the real world, including 74 training and 15 testing samples. Three different LDR images
in a sample are captured with exposure biases of $\{$-2, 0, +2$\}$ or $\{$-3, 0, +3$\}$. Hu’s dataset is a sensor-realistic synthetic dataset from a game engine, which is captured at three exposure levels (\ie,  $\{$-2, 0, +2$\}$). We use the dynamic scene images in Hu’s dataset. 
Following \cite{Hu2020Sensor}, we choose the first 85 samples for training, and the remainder 15 samples for testing. To verify generalization performance, we evaluate all methods on Sen’s dataset \cite{Sen2012Robust} and Tursun’s dataset \cite{Tursun2016An} (without ground truth). 


\textbf{Evaluation Metrics.}
We calculate five common metrics used for testing, \ie, PSNR-L, PSRN-$\mu$, SSIM-L, SSIM-$\mu$ and HDR-VDP-2 \cite{Mantiuk2011HDR}, where ‘-L’ denotes linear domain, ‘-$\mu$’ represents tonemapping domain.

\textbf{Implementation Details.}
We apply 8 $\times$ 8 patch size in the PA module. The layer numbers of STL and RDTB are 6 and 3. 
The window size of WDTL is 8 $\times$ 8. In the training stage, we crop the 128 $\times$ 128 patches with stride 64 for the training dataset. We use Adam optimizer, and set the batch size and learning rate as 4 and 0.0002, receptively. We drop the learning rate by 0.1 every 50 epochs. And we set $\beta_{1}$=0.9, $\beta_{2}$=0.999 and $\epsilon$=$1e^{-8}$ in Adam optimizer. We implement our model using PyTorch with 2 NVIDIA GeForce 3090 GPUs and train for 150 epochs.

\begin{table*}
\small
\caption{The evaluation results on Kalantari's \cite{Kalantari2017Deep} and Hu's \cite{Hu2020Sensor} datasets. The best and the second best results are highlighted in \textbf{Bold} and \underline{Underline}, respectively.}
\centering
\setlength{\tabcolsep}{0.96mm}
\label{table1}
\begin{tabular}{c|c|ccccccccccccc}
\noalign{\smallskip} \hline \noalign{\smallskip}
\textbf{Datasets} & \textbf{Models} & \textbf{Sen} & \textbf{Hu} & \textbf{Kalantari} & \textbf{DeepHDR} & \textbf{AHDRNet} & \textbf{NHDRR} & \textbf{HDR-GAN} & \textbf{ADNet} & \textbf{APNT} & \textbf{CA-ViT} & \textbf{Ours} \\
\hline

\specialrule{0em}{1pt}{1pt}
\specialrule{0em}{1pt}{1pt}

\multirow{5}{*}{Kalantari} & PSNR-$\mu$ & 40.95 & 32.19 & 42.74 & 41.64 & 43.62 & 42.41 & 43.92 & 43.76 & 43.94 & \underline{44.32} &\textbf{44.64} \\
                           & PSNR-L & 38.31 & 30.84 & 41.22 & 40.91 & 41.03 & 41.08 & 41.57 & 41.27 & 41.61 & \underline{42.18}
                           &\textbf{42.47} \\
                           & SSIM-$\mu$ & 0.9805 & 0.9716 & 0.9877 & 0.9869 & 0.9900 & 0.9887 & 0.9905 & 0.9904 & 0.9898
                           &\textbf{0.9916}
                           &\underline{0.9915} \\
                           & SSIM-L & 0.9726 & 0.9506 & 0.9848 & 0.9858 & 0.9862 & 0.9861 & 0.9865 & 0.9860 & 0.9879
                           &\underline{0.9884}
                           &\textbf{0.9894} \\
                           & HDR-VDP-2 & 55.72 & 55.25 & 60.51 & 60.50 & 62.30 & 61.21 & 65.45 & 62.61 & 64.05 
                           &\underline{66.03}
                           &\textbf{66.05} \\
\specialrule{0em}{1pt}{1pt}
\hline
\specialrule{0em}{1pt}{1pt}
\specialrule{0em}{1pt}{1pt}

\multirow{5}{*}{Hu} & PSNR-$\mu$ & 31.48 & 36.56 & 41.60 & 41.13 & 45.76 & 45.15 & 45.86 & 46.79 & 46.41 &\underline{48.10} &\textbf{48.46} \\
                    & PSNR-L & 33.58 & 36.94 & 43.76 & 41.20 & 49.22 & 48.75 & 49.14 & 50.38 & 47.97 
                    &\underline{51.17}
                    &\textbf{51.91} \\
                    & SSIM-$\mu$ & 0.9531 & 0.9824 & 0.9914 & 0.9870 & \underline{0.9956} & 0.9945 & \underline{0.9956} & 0.9908 & 0.9953 
                    & 0.9947 &\textbf{0.9959} \\
                    & SSIM-L & 0.9634 & 0.9877 & 0.9938 & 0.9941 & 0.9980 & \underline{0.9989} & 0.9981 & 0.9987 & 0.9986 &\underline{0.9989}
                    &\textbf{0.9991} \\
                    & HDR-VDP-2 & 66.39 & 67.58 & 72.94 & 70.82 & 75.04 & 74.86 & 75.19 & 76.21 & 73.06
                    &\underline{77.12}
                    &\textbf{77.24} \\  

\noalign{\smallskip} \hline \noalign{\smallskip}
\end{tabular}
\end{table*}

\subsection{Comparison with the State-of-the-art Methods}
To evaluate our model, we perform quantitative and qualitative experiments comparing with several state-of-the-art deep learning-based methods: Kalantari's method \cite{Kalantari2017Deep}, DeepHDR \cite{Wu2018Deep}, AHDRNet \cite{Yan2019Attention}, NHDRR \cite{Yan2020Deep}, HDR-GAN \cite{Niu2021Hdr}, ADNet \cite{Liu2021ADNet}, APNT \cite{Chen2022Attention}, CA-ViT\cite{liu2022ghost}. 

\subsubsection{Datasets w/ Ground Truth.}

As shown in Figure \ref{fig:KalanHuTursunSen} (a) and (b), these two datasets have some challenging samples that cover a large area of foreground motions and over/under-exposed regions. 
Due to intractable large motion and occlusion, most comparing approaches produce ghosting artifacts in these areas. Kalantari’s method and DeepHDR cannot handle the motion of the background due to error-prone alignments (\ie, optical flow and tomographies), which cause undesirable ghosting (See the red block in Figure \ref{fig:KalanHuTursunSen} (a)(b)). Since NHDRR and HDR-GAN do not have explicit alignment, they cannot recover the details in the saturated areas, therefore generating color distortions. Although AHDRNet and ADNet alleviate ghosting artifacts using ghost attention, they also suppress some useful neighborhood information, and cannot reconstruct large motion overlapped with over/under-exposure cases (See the blue block in Figure \ref{fig:KalanHuTursunSen} (a) and (b)). APNT generates ghost regions and blurred edges because of the limitation of the patch-matching strategy. Without patch-level alignment, CA-ViT also produces ghosts. 
(See the red block in Figure \ref{fig:KalanHuTursunSen} (a) and blue block in Figure \ref{fig:KalanHuTursunSen} (b)).
Thanks to the proposed patch aggregation and gating module, the proposed HyHDRNet not only integrates several similar patch features to complete misaligned or saturated regions, but also recovers sharp edges in motion or saturation.

The quantitative results for the proposed HyHDRNet on two datasets are shown in Table \ref{table1}. The proposed HyHDRNet achieves state-of-the-art performance consistently on all five metrics of two datasets. We show that the improvement of HyHDRNet is very obvious against other previous SOTA methods. HyHDRNet surpasses the second-best methods by 0.32 db, 0.29 db in terms of PSNR-$\mu$ and PSNR-L and on Kalantari’s dataset \cite{Kalantari2017Deep}, and it also improves by 0.36 db and 0.74 db in terms of PSNR-$\mu$ and PSNR-L on Hu’s dataset \cite{Hu2020Sensor}.


\subsubsection{Evaluation on Datasets w/o Ground Truth.}
We compare the proposed HyHDRNet with other approaches on Tursun’s \cite{Tursun2016An} and Sen’s \cite{Sen2012Robust} datasets, which do not have ground truth. Visual comparisons are shown in Figure \ref{fig:KalanHuTursunSen} (c)(d). As can be seen, most methods cannot recover the large area of saturation region and large motion. In Figure \ref{fig:KalanHuTursunSen} (c) (red zoomed-in block), the halo effect of the sun is evident in the results of other methods, causing saturated ghosting artifacts. The over-exposure problem also exists in Figure \ref{fig:KalanHuTursunSen} (d). Thanks to the proposed patch aggregation module and deformable transformer, HyHDRNet can assemble similar patches to fill up the missed content and adaptively fuse useful texture in a larger receptive field (See the red block in Figure \ref{fig:KalanHuTursunSen} (c)).
For large motion reconstruction problems, other methods have ghosting artifacts of moving car in the blue block of Figure \ref{fig:KalanHuTursunSen} (c). Because the gating module can leverage both patch-wise aggregation and pixel-wise ghost attention, HyHDRNet can generate pleasing ghost-free results. 


\begin{table}
\small
\caption{The Ablation study on Kalantari dataset}
\centering
\setlength{\tabcolsep}{1.1mm}
\label{table2}
\begin{tabular}{l|ccc}
\noalign{\smallskip} \hline \noalign{\smallskip}
\textbf{Models} & \textbf{PSNR-$\mu$} & \textbf{PSNR-L} & \textbf{HDR-VDP-2} \\
\hline
\specialrule{0em}{1pt}{1pt}
\specialrule{0em}{1pt}{1pt}

1.Baseline & 43.56 & 41.72 & 63.80  \\
2.+GA & 43.99 & 41.95 & 65.43  \\
3.+PA & 44.22 & 42.03 & 65.57  \\
4.+GA+PA+Gating & 44.49 & 42.32 & 65.98  \\
5.+GA+PA+Addition & 44.28 & 42.10 & 65.79 \\
6.+GA+PA+Concat & 44.32 & 42.16 & 65.84 \\
7.+GA+PA+Gating+DL & 44.60 & 42.41 & 66.03 \\
8.A-DRDB & 44.06 & 41.66 & 65.49  \\
9.Ours & \textbf{44.64} & \textbf{42.47} & \textbf{66.05}  \\

\noalign{\smallskip} \hline \noalign{\smallskip}
\end{tabular}
\end{table}

\subsection{Ablation Studies}
We conduct ablation studies on Kalantari's dataset, and validate the effectiveness of each proposed module in HyHDRNet. We use the following variants of HyHDRNet:
1) \textbf{Model1}: As a baseline, we concatenate three images as input to the fusion network that employs the same RSTB setting in SwinIR \cite{Liang2021Swinir}. 2) \textbf{Model2}: We add GA module into model1. 3) \textbf{Model3}: PA module is added into model1. 4) \textbf{Model4}: We integrate GA, PA and gating modules into model1. 5) \textbf{Model5}: We replace gating with addition. 6) \textbf{Model6}: We replace gating with concatenation. 7) \textbf{Model7}: We add Deformable Layer (DL) into model4. 8)  \textbf{Model8}: We replace Transformer-based fusion subnetwork with DRDB convolution network \cite{Yan2019Attention}, named Alignment-DRDB (A-DRDB).
9) \textbf{Model9}: The proposed HyHDRNet full model, which adds perceptual loss into model7.\par

\begin{figure}
    \centering
    \includegraphics[width=0.4\textwidth]{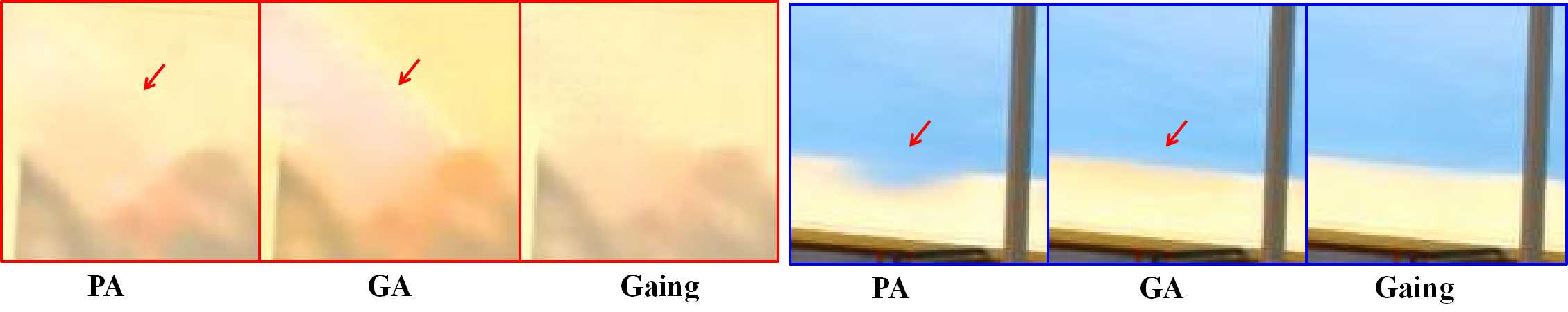}
    \caption{Visual results of ablation study.}
    \vspace{-0.2cm}
    \label{fig:ablation}
\end{figure}

\begin{figure}
  \centering
  \includegraphics[width=0.25\textwidth]{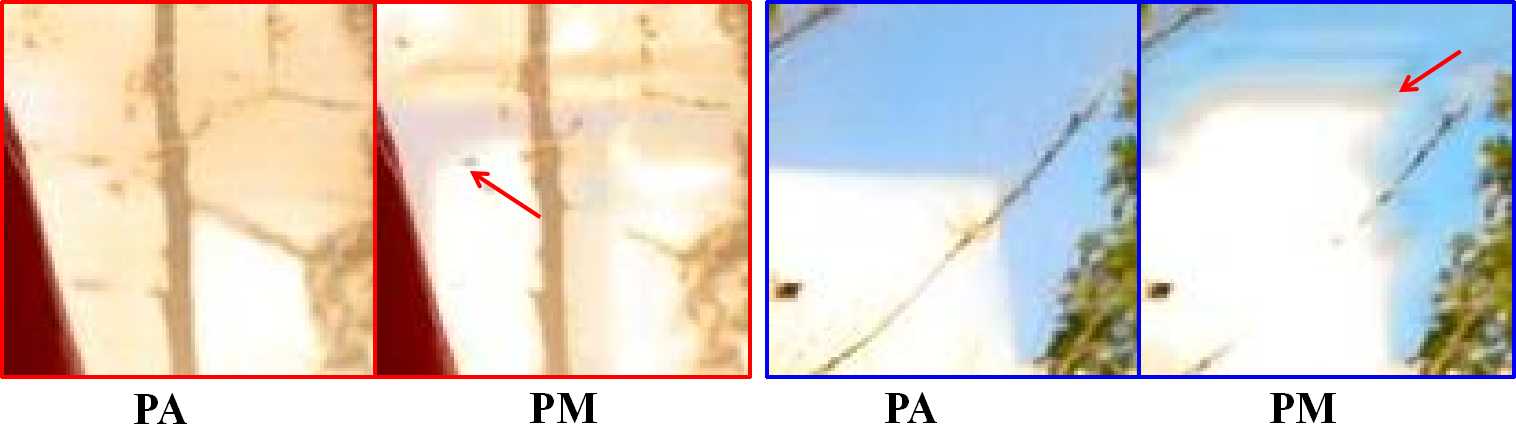}
  \caption{Visual results on PM and PA methods.}
  \label{fig:supp_ablation2}
  \vspace{-0.3cm}
\end{figure}

\textbf{PA and GA modules.} As shown in Table \ref{table2}, compared with Model1, the performance of Model2 and Model3 are obviously improved. It demonstrates that the proposed PA and GA modules are both effective mechanisms for ghost removal. Note that Model3 with PA module achieves better numerical results than Model2, which means that the PA module is a better method for obtaining contents. 
As shown in Figure \ref{fig:ablation}, the red block results of PA (Model3) do not have ghosting artifacts, which verifies that the PA module can selectively aggregate useful information from non-reference LDR images to remove ghosts and generate content for saturation. However, the PA module will cause blurry edges (see blue block results of PA).
Compared with patch aggregation, the results of Model2 (GA) have sharp edges (blue block of Figure \ref{fig:ablation}) but cannot remove ghosts completely. We consider it can be attributed to ghost attention suppress undesired components with pixel level.
\textbf{Gating module.} Since the gating module can realize mutual guidance of PA and GA modules, the numerical (Table \ref{table2}) and visual (Figure \ref{fig:ablation}) results are both improved. 
As shown in Figure \ref{fig:ablation}, since Model4 with GA can achieve mutual guidance of PA and GA modules in the content alignment subnetwork, the results can remove ghosts and hold on sharp edges (see last column), simultaneously.
In addition, we also employ Addition (Model5) and Concatenation (Model6) to replace gating, but performance is degraded (See Table \ref{table2}), which validates the effectiveness of the gating module.
\textbf{Transformer-based Fusion Subnetwork.} In Table \ref{table2}, compared Model7 with Model4, deformable layer (DL) obtains a better result. It can be attributed to the advantage of DL which captures more information for the fusion subnetwork.  
When we replace the Transformer-based Fusion Subnetwork with DRDBs, the numerical results obviously decrease, which demonstrates the effectiveness of the Transformer-based Fusion Subnetwork. 
\textbf{Patch Aggregation vs Patch Matching.} To verify the effectiveness of the patch aggregate module further, we show the qualitative and quantitative results of patch aggregation and patch matching in Figure \ref{fig:supp_ablation2} and Table \ref{table3}. While the traditional patch matching-based HDR imaging methods have excellent performance, these methods only choose one best patch to reuse the original patches, this operation has a low patch utilization ratio which causes insufficient content during fusion.
As shown in Figure \ref{fig:supp_ablation2}, since saturated regions are hard to recover content with only one patch, thus the results still have obvious saturation regions.
Different from patch matching, patch aggregation can selectively aggregate useful information from non-reference LDR images, which can generate content for saturation.

\begin{table}
\small
\caption{The Ablation study on Patch Utilization method}
\centering
\setlength{\tabcolsep}{3pt}
\label{table3}
\begin{tabular}{c|ccc}
\noalign{\smallskip} \hline \noalign{\smallskip}
\textbf{Patch methods}  &\textbf{PSNR-$\mu$} &  \textbf{PSNR-L} & \textbf{HDR-VDP-2} \\
\hline
\specialrule{0em}{1pt}{1pt}
\specialrule{0em}{1pt}{1pt}

PM  & 43.24 & 40.60 & 63.10 \\
PA  & \textbf{44.64} & \textbf{42.47} & \textbf{66.05} \\

\noalign{\smallskip} \hline \noalign{\smallskip}
\end{tabular}
\end{table}


\vspace{-0.2cm}

\section{Conclusion}
We propose an effective HDR imaging method based on content alignment and Transformer-based fusion. In content alignment subnetwork, we employ patch aggregation module to selectively aggregate useful patches from non-reference LDR images. We also propose a novel window-based deformable Transformer block to fuse a large range of information from extracted features. The major advantage of our proposed method is that it can fuse similar content from non-reference LDR images to remove ghosts and generate content for saturation in complex cases. Further, we demonstrate the superiority of our method over existing state-of-the-art methods on four publicly available datasets.


{\small
\bibliographystyle{ieee_fullname}
\bibliography{egbib}
}

\end{document}